\PassOptionsToPackage{dvipsnames}{xcolor}
\documentclass[11pt,letterpaper]{style}

\usepackage[numbers]{natbib}
\usepackage{graphicx}
\usepackage{booktabs}
\usepackage{amsmath,amsfonts,amssymb}
\usepackage{cleveref}
\usepackage{subcaption}
\usepackage{wrapfig}
\usepackage{multirow}
\usepackage{colortbl}
\usepackage{listings}
\usepackage{xparse}
\usepackage{fontawesome5}
\usepackage{bxcoloremoji}
\usepackage{float}
\usepackage{placeins}
\usepackage{threeparttable}

\graphicspath{{./}{fig/}{figures/}{plot/}{pdf/}{table/}}
\usepackage{amsthm}
\usepackage{tcolorbox}
\usepackage{svg}
\tcbuselibrary{skins,breakable}
\tcbuselibrary{listingsutf8}
\usepackage{titletoc}

\usepackage{setspace}
\usepackage{pifont}
\usepackage{mathtools}
\usepackage{enumitem}
\usepackage{arydshln}
\usepackage{bbm}
\usepackage{lineno}
\usepackage{makecell}
\usepackage{adjustbox}
\usepackage{algorithm}
\usepackage{algorithmic}
\usepackage{caption} 
\usepackage{wrapfig}
\usepackage{hyperref}
\usepackage{bbding}
\usepackage{xspace}

\hypersetup{colorlinks=true,linkcolor=red,urlcolor=blue,citecolor={blue}}

\definecolor{mygray}{gray}{0.9}
\definecolor{syncol}{RGB}{243,246,249}
\definecolor{wildcol}{RGB}{215,240,235}
\definecolor{drop1}{RGB}{180,225,220}
\definecolor{drop2}{RGB}{150,210,200}
\definecolor{drop3}{RGB}{120,195,185}
\definecolor{drop4}{RGB}{95,180,170}
\definecolor{drop5}{RGB}{65,160,150}
\definecolor{lightblue}{RGB}{26,82,249}

\definecolor{myblue1}{HTML}{0171DC}
\definecolor{myblue2}{HTML}{013978}

\NewDocumentEnvironment{minted}{O{} m +b}{%
}{}

\newcommand{\equal}{\textsuperscript{*}}     

\renewcommand\Authfont{\centering\normalfont\bfseries\fontsize{11}{15}\selectfont}
\renewcommand\Affilfont{\centering\normalfont\fontsize{10}{15}\selectfont}

\title{MASkills: Continual Skills Optimization for Multi-Agent LLM Systems}
\runningtitle{MASkills: Continual Skills Optimization for Multi-Agent LLM Systems}

\author{%
    {\Authfont
    \textbf{Huaiyuan Yao}\textsuperscript{1} \quad
    \textbf{Xiaoou Liu}\textsuperscript{1} \quad
    \textbf{Charles Fleming}\textsuperscript{2} \quad
    \textbf{Tianlong Chen} \textsuperscript{3} \quad
    \textbf{Hua Wei
 }\textsuperscript{1}
    }\\
    {\Affilfont
    \textsuperscript{1}  Arizona State University \quad \texttt{\{huaiyuan, xiaoouli, hua.wei\}@asu.edu} \\
   \textsuperscript{2}  Cisco Research \quad 
     \texttt{chflemin@cisco.com} \\
    \textsuperscript{3}  University of North Carolina at Chapel Hill \quad
     \texttt{tianlong@cs.unc.edu}
    }
}

\begin{document}
\begin{abstract}
LLM-based multi-agent systems have shown strong performance on complex tasks, yet continual improvement from interaction experience remains challenging. Existing self-reflection methods build experience memories, but memories are mostly hard to invoke, refine, or scale, while agent skills offer a more actionable unit: structured procedural knowledge that specifies when to act, how to act, and which resources or tools to use. We introduce \textbf{MASkills}, a continual learning framework that optimizes multi-agent LLM systems through agent skills. MASkills presents a new agent-optimization pipeline that integrates skill-conditioned credit assignment, hierarchical credit aggregation, and momentum-smoothed optimization, enabling agent skill libraries to evolve through refinement, induction, consolidation, and pruning. Experiments on HotpotQA, LoCoMo, and GAIA demonstrate the effectiveness of MASkills across multiple agentic tasks. Our code is available at \url{https://github.com/DaRL-GenAI/MASkills}
\end{abstract}
\newcommand{\TitleLinks}{%
\centering
    \vspace{8pt}
}
\maketitle

\section{Introduction}
\label{sec:intro}

LLM-based multi-agent systems have emerged as a practical paradigm for solving complex tasks through coordination, role specialization, and long-horizon interaction~\cite{liu2026androidreality, yao2025comal, zhao2025fuas}. However, how to continually improve a multi-agent LLM system over time remains challenging. Existing methods typically use self-reflection to construct an experience memory~\cite{tsui2025selfcorrection, yao2025lilodriver}. Such memories can preserve useful experience, but they remain a weak basis for continual improvement: they record what happened in past trajectories, but not which action policy should be reused; they lack reliable invocation conditions; and as the memory grows, useful lessons become mixed with noisy, redundant, or stale free-form records~\cite{zhang2024surveymemorymechanismlarge}.

A more useful unit for continual improvement is a \textit{skill}. According to Anthropic~\cite{anthropic2025skills}, a skill is a structured package of procedural knowledge that tells an agent how to perform a class of tasks, including when to invoke it, how to act, which resources or tools to use. Skills make experience reusable: agents can discover relevant skills from lightweight descriptions, load detailed instructions and resources on demand, and refine them as new experience accumulates. This progressive-disclosure design makes skills more scalable and actionable than large unstructured memory stores. Recent methods~\cite{memskill2026, evoskill2026, polyskill2026} develop such reusable procedural abstractions, but remain predominantly single-agent and do not optimize skills under the coordination dynamics that determine team-level utility in multi-agent systems.



Extending skill evolution to multi-agent settings requires agents to refine their skills through interaction, so that skill updates reflect not only individual outcomes but also their contributions to team-level coordination. This goal raises three concrete challenges: \textbf{(1) Skill-level credit assignment.} Team rewards should be attributed not only to the correct agent, but also to the correct skill. In multi-agent settings, a skill may only be invoked at a few timesteps, and its effect is entangled with the behaviors and skills of other agents. As a result, team-level or even agent-level feedback is insufficient to determine which specific skills should be improved. \textbf{(2) Noisy and heterogeneous interaction.} Language-based feedback extracted from trajectories is often noisy, inconsistent, and unstable across learning cycles. Moreover, multi-agent systems may operate under very different coordination structures, such as centralized planning or decentralized peer communication, which induce different patterns of interaction dependency and credit propagation. \textbf{(3) Discrete and high-risk skill updates.} Unlike parameter vectors, skills are discrete language artifacts such as \texttt{SKILL.md} files, scripts, and references. Skill evolution, therefore, requires open-ended structural modifications rather than smooth numerical updates. Incorrect edits may directly alter the agent's effective action space, making continual skill optimization inherently high-risk and difficult to stabilize.

To address these challenges, we introduce \textbf{MASkills}, a language-based continual learning framework that performs policy improvement directly in agents’ skill spaces rather than parameter spaces. Under decentralized execution, each agent invokes reusable skill artifacts as part of its latent policy reasoning, while the system continually evolves these skills through language-space optimization. MASkills adopts a unified language-space policy-gradient-style framework, following the structural interpretation of TextGrad~\cite{yuksekgonul2024textgrad} and LangMARL~\cite{langmarl2026}, with three components corresponding to the challenges above: (1) skill-conditioned credit assignment, where a critic attributes trajectory feedback to specific skill invocations through counterfactual comparison; (2) hierarchical credit aggregation and momentum-smoothed optimization, which stabilize noisy feedback across trajectories, agents, skills, and interaction topologies; and (3) credit-driven skill-space optimization, where stable credit signals drive skill refinement, induction, consolidation, and pruning, together with held-out validation and rollback mechanisms.

Our contributions are summarized as follows: (1) We introduce a new paradigm for continual multi-agent LLM learning that casts policy improvement as optimization over reusable skill spaces distilled from interaction traces. (2) We propose \textbf{MASkills}, a continual learning framework for multi-agent skill optimization that combines skill-conditioned credit assignment, aggregation, and optimization to evolve agent skill libraries through refinement, induction, consolidation, and pruning, together with held-out validation and rollback for stable updates. (3) We evaluate MASkills on HotpotQA, LoCoMo, and GAIA, demonstrating strong performance across memory and agentic tasks.

\section{Related Work}
\label{sec:related}

\paragraph{Agent Skill Discovery.}
\textsc{Voyager}~\cite{wang2023voyager} pioneered an ever-growing skill library for an embodied LLM agent. 
Recent work has extended this paradigm in three complementary directions.
\textsc{MemSkill}~\cite{memskill2026} treats memory operations themselves as a learnable skill bank: a controller selects a Top-$K$ subset of skills per text span, an LLM executor applies them, and a designer periodically revises the bank from clustered hard cases. \textsc{EvoSkill}~\cite{evoskill2026} casts skill discovery as iterative textual feedback descent: an Executor runs tasks, a Proposer diagnoses failures, and a Skill-Builder materializes proposals into structured \texttt{SKILL.md} folders, retained on a Pareto frontier only if they improve held-out validation. \textsc{PolySkill}~\cite{polyskill2026} introduces polymorphic abstraction, separating a skill's abstract goal from its concrete site-specific implementation, enabling cross-website transfer. However, these methods assume single-agent feedback and cannot directly attribute team-level outcomes to interacting agents and skills.

\paragraph{Multi-agent LLM Systems.}
LLM multi-agent frameworks such as AutoGen~\cite{wu2024autogen}, MetaGPT~\cite{hong2023metagpt}, agentic neural networks~\cite{ma2025agentnet}, and symbolic learning~\cite{ou2025symbolic} have demonstrated that division of labor among LLMs improves coverage on complex tasks. Most of these systems rely on hand-engineered roles and prompts~\cite{yao2026instructional, chen2026every, hu2026small}; the few that self-improve do so by either rewriting prompts globally such as TextGrad~\cite{yuksekgonul2024textgrad}, DSPy~\cite{khattab2023dspy} or sharing a memory like Reflexion~\cite{shinn2023reflexion}. LangMARL~\cite{langmarl2026} introduced explicit credit assignment into language-space optimization.



\section{Problem Formulation}
\label{sec:prelim}

We formalize MASkills on top of a cooperative multi-agent setting, a skill artifact, and a language-based notion of credit.

\paragraph{Cooperative multi-agent setting.} We model a team of $N$ LLM agents as a Dec-POMDP
\begin{equation}
\mathcal{M}=\langle S, \{A_i\}_{i=1}^N, \{O_i\}_{i=1}^N, P, R, \gamma \rangle
\end{equation}
The agents share a team reward $r_t = R(s_t, a_t^1, \dots, a_t^N)$, and each agent $i$ executes a decentralized policy $\pi_i(a_t^i \mid o_t^i)$ that maps its local observation to an action. The team objective is the expected discounted return $J(\pi)=\mathbb{E}_\pi[\sum_t \gamma^t r_t]$. MASkills inherits this Dec-POMDP and re-parameterizes each $\pi_i$ through a skill library.

\paragraph{Skill artifacts.}
Following~\cite{evoskill2026}, we model each skill artifact as a structured, file-system-based representation. For agent \(i\), the \(j\)-th skill artifact is defined as
\begin{equation}
k_i^{(j)} = \left(y_i^{(j)}, m_i^{(j)}, \mathcal{R}_i^{(j)}\right),
\end{equation}
where \(y_i^{(j)}\) denotes the metadata file, such as \texttt{skill.yaml}; 
\(m_i^{(j)}\) denotes the procedural instruction file, such as \texttt{SKILL.md}; 
and \(\mathcal{R}_i^{(j)}\) denotes a set of auxiliary resources, such as scripts, references, assets, and configuration files.
In implementations such as Claude/Codex-style skills, this representation often corresponds to a directory containing a procedural instruction file and optional resource folders, while some variants separate metadata into a dedicated \texttt{skill.yaml} file. This structure supports concise procedural abstraction, actionable execution guidance, hierarchical disclosure, and reusable skill composition.

Each agent \(i\) maintains an evolving \emph{skill library} $\mathcal{K}_i = \{k_i^{(1)}, k_i^{(2)}, \dots, k_i^{(M_i)}\},$
where \(M_i\) denotes the current number of skills available to agent \(i\). We treat \(\mathcal{K}_i\) as the learnable component of agent \(i\)'s policy. MASkills therefore optimizes reusable structured skill artifacts that parameterize decentralized agent behavior.

\paragraph{Skill-space optimization.}
Whereas classical MARL optimizes network parameters $\theta$, MASkills directly optimizes the agents' skill libraries:
\begin{equation}
\max_{\{\mathcal{K}_i\}_{i=1}^N}
\ \mathbb{E}_{\tau \sim \pi(\cdot \mid \{\mathcal{K}_i\})}
\bigl[R(\tau)\bigr].
\label{eq:skill_obj}
\end{equation}
Here, the optimization variables $\{\mathcal{K}_i\}$ are skill artifacts. We approach Eq.~\ref{eq:skill_obj} through a \textit{language-space policy-gradient-style} framework, following the structural interpretation introduced by TextGrad~\cite{yuksekgonul2024textgrad}. The analogy is conceptual rather than differentiable: decentralized rollouts provide trajectory sampling, language critics provide skill-level advantage-like feedback, hierarchical aggregation and momentum smoothing approximate Monte-Carlo expectation and variance reduction, and credit-driven skill editing operators act as pseudo-gradient updates over each $\mathcal{K}_i$. Throughout the paper, terms such as \textit{gradient}, \textit{advantage}, and \textit{momentum} should therefore be understood as language-space analogies rather than literal derivatives.

\begin{figure*}[t]
  \centering
  \includegraphics[width=\textwidth]{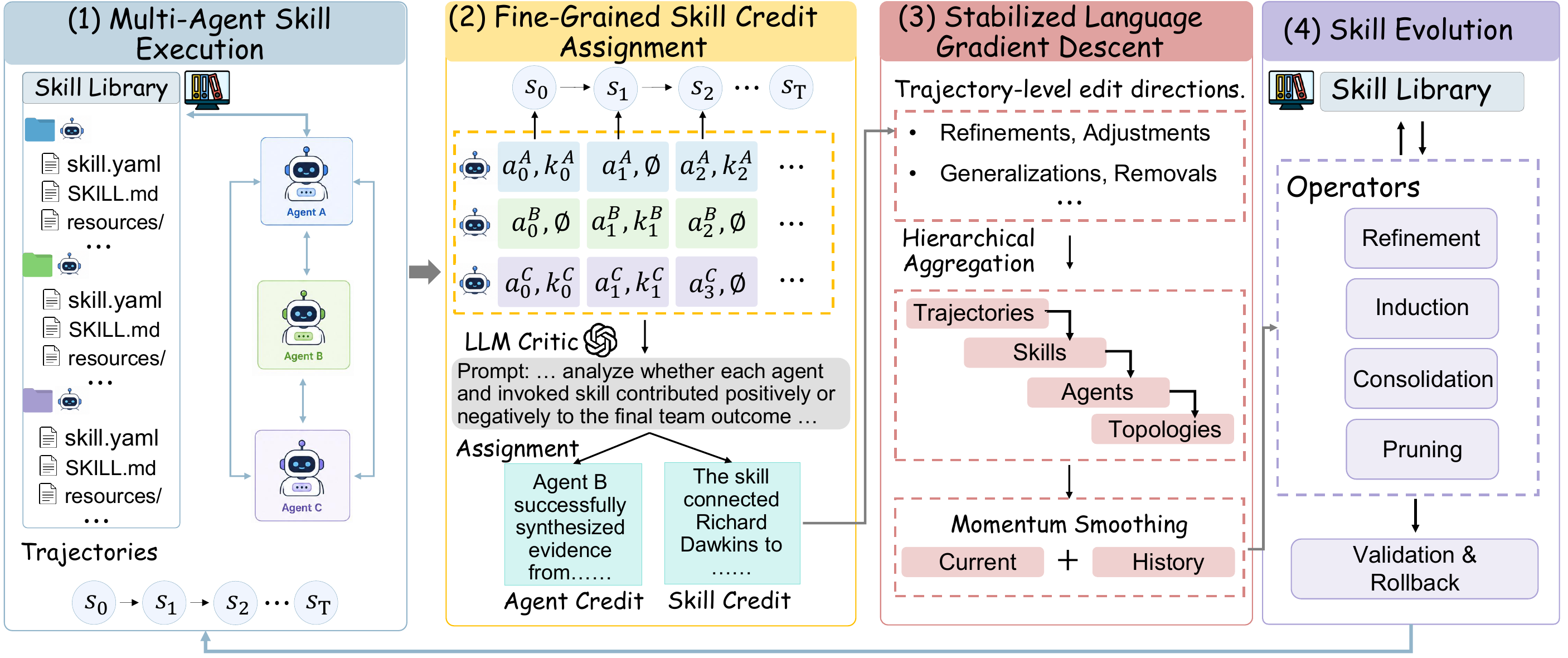}
  \caption{Overview of the MASkills pipeline as a four-step language-space analog of policy optimization. Agents execute skills to generate trajectories, receive skill-conditioned linguistic credit, aggregate temporally stable advantages, and update the skill library through refinement, induction, consolidation, and pruning.}
\label{fig:system}
\end{figure*}
\section{MASkills Framework}
\label{sec:method}

MASkills operationalizes the skill-space objective as a closed-loop process structured as a language-space analogue of a policy gradient: decentralized rollouts produce skill traces; a language critic emits per-agent, per-skill counterfactual credits; a hierarchical aggregator over a batch plus momentum smoothing across cycles produces a temporally stable skill advantage; and this advantage drives four credit-driven operators on each skill space under a held-out validation, as shown in Fig.~\ref{fig:system}.

\subsection{Multi-Agent Skill Execution}

Each agent is an LLM equipped with a skill space
$\mathcal{K}_i$.
At rollout time, lightweight metadata from each
\texttt{skill.yaml} (e.g., skill name and short description) is injected into
the agent prompt, exposing the currently available skills to the model.
The agent then autonomously decides whether to invoke a skill, which skill to
invoke, and how to compose multiple skills during reasoning.

Skills are exposed as callable tools rather than fully expanded prompts.
When a skill is selected, its corresponding \texttt{SKILL.md} and auxiliary
resources $\mathcal{R}$ are dynamically loaded into the context. This keeps the active context compact while enabling specialized
procedural behavior. Formally, the agent policy is conditioned on its local observation $o_t^i$, interaction history $h_t^i$, and skill space $\mathcal{K}_i$:
\begin{equation}
a_t^i
\sim
\pi_\theta\bigl(
\cdot \mid o_t^i,\mathcal{K}_i,h_t^i
\bigr).
\label{eq:skill_policy}
\end{equation}

Let $ c_t^i \in \mathcal{K}_i \cup \{\varnothing\} $
denote the skill invoked by agent $i$ at timestep $t$, where
$c_t^i=\varnothing$ means the agent acted directly without loading a skill.
Since skill invocations are explicit tool calls, the resulting skill trace
$ \xi_i = \{(t,c_t^i)\}_{t=0}^{T-1} $
is directly observable from execution logs.

The resulting trajectory becomes
$
\tau
=
\bigl\langle
s_0,
\{a_0^i,\xi_0^i\}_{i=1}^N,
\dots,
s_T
\bigr\rangle,
$
augmenting the standard multi-agent trajectory with explicit skill-invocation
records that later support skill-level credit assignment.

\subsection{Fine-Grained Skill Credit Assignment}
\label{ssec:credit}

Optimizing LLM agents at the skill level requires identifying which invoked
skills actually contributed to team success or failure.
Agent-level feedback alone is insufficient: a single trajectory may involve
multiple interacting skills across multiple agents, and only a subset of them
may be responsible for the final outcome.

MASkills therefore assigns credit at the granularity of individual skill
invocations.
Given a trajectory $\tau$ with skill traces, a
centralized language critic evaluates how each invoked skill affected the team
outcome relative to the counterfactual where that skill had not been used.

For each skill
$k \in \mathcal{K}_i(\tau)$
actually invoked by agent $i$ along trajectory $\tau$, the critic produces a
structured natural-language credit:
\begin{equation}
C_i^{\text{text}}(\tau,k)
=
\mathrm{LLM}_{\text{Critic}}
(\tau,i,k,\xi_i),
\label{eq:skill_credit}
\end{equation}
which explains whether the skill helped coordination, was redundant, caused a
failure, or should be generalized or specialized. In addition to per-skill feedback, the critic also emits an agent-level
residual credit:
$
C_i^{\text{text}}(\tau)
=
\mathrm{LLM}_{\text{Critic}}(\tau,i),
$
which captures effects not attributable to any single skill.
This residual is particularly important for skill induction: if a failure
cannot be explained by existing skills, the agent-level critique signals that
a new skill may be needed.

\subsection{Stabilized Language Gradient Descent}
\label{ssec:agg}

Trajectory-level language critiques are inherently noisy and often contradictory across rollouts. To stabilize optimization, MASkills converts raw critiques into stable language-space gradients through gradient extraction, hierarchical aggregation, and momentum-based updates.

\paragraph{Trajectory-level edit directions.}

Given a trajectory $\tau$ and an invoked skill $k$,
the language critic first converts the textual credit signal into a
trajectory-level edit direction:
\begin{equation}
g_i(\tau,k)
=
\mathrm{LLM}_{\text{Grad}}
\Bigl(
C_i^{\text{text}}(\tau,k)
\Bigr),
\label{eq:grad}
\end{equation}
where $\mathrm{LLM}_{\text{Grad}}(\cdot)$ extracts a structured update direction from the credit signal, including suggested refinements, behavioral adjustments, generalizations, or removals in skill space. Unlike numerical gradients, $g_i(\tau,k)$ is a structured natural-language edit direction over skills.

\paragraph{Hierarchical aggregation.} Given a rollout batch $\mathcal{B}=\{\tau_1,\dots,\tau_B\}$, trajectory-level gradients are recursively aggregated across trajectories, skills, agents, and interaction topologies. At the skill level, the gradients associated with skill $k$ are merged into a single language-space gradient estimate:
\begin{equation} G_i^{(m)}(k) = \mathrm{LLM}_{\text{Agg}} \Bigl( \{ g_i(\tau_b,k) \}_{b=1}^B \Bigr), \label{eq:agg} \end{equation}
where $\mathrm{LLM}_{\text{Agg}}(\cdot)$ is an LLM-based operator that merges recurring behavioral patterns, resolves conflicting update directions, removes redundancy, and summarizes the coordination utility of the skill. Higher-level aggregation recursively combines these gradients across agents into a stable language gradient estimate.

\paragraph{Momentum-stabilized skill editing.}

Even after aggregation, edit directions may still fluctuate across
optimization cycles.
Instead of treating each cycle independently,
MASkills carries forward persistent historical edit directions together with
the current aggregated feedback.

The skill-space update therefore follows a language-space analogue of
momentum gradient descent:
\begin{equation}
\mathcal{K}_i^{(m+1)}
=
\mathrm{LLM}_{\text{Edit}}
\Bigl(
\mathcal{K}_i^{(m)},
\,
G_i^{(m)},
\,
G_i^{(m-1)}
\Bigr),
\label{eq:skill_update}
\end{equation}
where $\mathrm{LLM}_{\text{Edit}}(\cdot)$ denotes a language-space editing procedure that modifies the skill library using both the current aggregated feedback and persistent historical edit directions from previous optimization cycles. Conceptually, the current edit direction plays the role of the current gradient update, while the historical edit direction acts as a momentum term that preserves persistent behavioral improvements over time. This stabilization suppresses transient or contradictory critiques and encourages consistent long-horizon skill evolution. In practice, the abstract editing operator $\mathrm{LLM}_{\text{Edit}}(\cdot)$ is instantiated through four concrete skill-space transformations: refinement, induction, consolidation, and pruning.

\subsection{Skill Evolution Operators}
\label{ssec:operators}

MASkills improves agent behavior by directly modifying the skill space
$\mathcal{K}_i$.
Since skills are structured language artifacts rather than differentiable
parameters, policy improvement is realized through a set of credit-driven
editing operators acting on different regions of the skill space.

\paragraph{Refinement.} Useful but imperfect skills are refined through localized diff-style edits guided by the aggregated edit direction: \begin{equation} k \leftarrow k \oplus \Delta k, \qquad \Delta k = \mathrm{LLM}_{\text{Refine}} \bigl( k, G_i^{(m)}(k) \bigr), \label{eq:refine} \end{equation} where $\Delta k$ is a structured edit patch rather than a full skill rewrite. The refinement operator modifies only the skill regions implicated by the aggregated feedback while preserving unrelated procedural structure and previously validated behavior.

\paragraph{Induction.}

When persistent failures or unresolved coordination patterns cannot be
adequately addressed by the current skill library,
MASkills expands the skill space by inducing new skills from difficult
trajectories. Let $\mathcal{H}_i$ denote the set of hard trajectories associated with
persistent unresolved critiques, missing behavioral capabilities,
or repeated coordination failures for agent $i$.
The system proposes a new skill
\begin{equation}
k_{\text{new}}
\sim
\mathrm{LLM}_{\text{Propose}}
\bigl(
\mathcal{H}_i,
G_i^{(m)}
\bigr),
\label{eq:induce}
\end{equation}
where the proposal is conditioned on both the hard-case trajectories and the
aggregated edit directions accumulated during optimization. Unlike refinement, which applies localized edits to existing skills,
induction introduces entirely new procedural abstractions into the skill space.

\paragraph{Consolidation.}

To prevent uncontrolled growth and fragmentation of the skill space,
MASkills periodically consolidates related skills into higher-level procedural
abstractions. Given a set of functionally overlapping skills
$\{k_a,k_b,\dots\}$,
it constructs a consolidated skill
\begin{equation}
k_{\text{macro}}
=
\mathrm{LLM}_{\text{Merge}}(k_a,k_b,\dots),
\label{eq:consolidate}
\end{equation}
where $\mathrm{LLM}_{\text{Merge}}(\cdot)$ synthesizes shared behavioral structure and reusable procedural patterns into a unified skill. Consolidation reduces redundancy in the skill library and encourages reusable higher-level abstractions.

\paragraph{Pruning.}
Not all skills remain useful as the skill space evolves.
Some skills become obsolete after consolidation,
others exhibit persistently poor coordination utility,
and some are rarely invoked across trajectories.
To control skill-space growth and remove low-value behaviors,
MASkills periodically prunes skills with consistently weak utility signals. Given the momentum-stabilized edit directions accumulated across optimization
cycles, low-utility skills are removed from the skill library:
\begin{equation}
\mathcal{K}_i
\leftarrow
\mathcal{K}_i
\setminus
\{
k :
\mathrm{LLM}_{\text{LowUtility}}
\bigl(
G_i^{(m)}(k)
\bigr)
\},
\label{eq:prune}
\end{equation}
where $\mathrm{LLM}_{\text{LowUtility}}(\cdot)$ identifies skills whose behavioral
contribution remains persistently negative, redundant, unstable,
or negligible across trajectories.

\paragraph{Validation and rollback.}

Because skill updates directly modify the agent's skill space,
incorrect edits may introduce behavioral regressions.
To ensure stable skill evolution,
all candidate edits are evaluated on a held-out validation set before being
committed to the skill library. Let
$\mathcal{D}_{\mathrm{val}}$
denote a held-out set of validation tasks and rollouts that are not used during
the current optimization cycle.
Given a candidate updated skill space
$\mathcal{K}_i'$,
MASkills estimates its validation performance while keeping all other agents'
skills fixed:
\begin{equation}
\hat{J}_{\mathrm{val}}
(\mathcal{K}_i' \mid \mathcal{K}_{-i})
=
\mathbb{E}_{\tau \sim \mathcal{D}_{\mathrm{val}}}
\bigl[
R(\tau)
\bigr].
\label{eq:validation}
\end{equation}

A candidate update is accepted only if
\begin{equation}
\hat{J}_{\mathrm{val}}
(\mathcal{K}_i' \mid \mathcal{K}_{-i})
\ge
\hat{J}_{\mathrm{val}}
(\mathcal{K}_i \mid \mathcal{K}_{-i})
-
\delta,
\label{eq:rollback}
\end{equation}
where $\delta \ge 0$ absorbs rollout variance and evaluation noise.
Otherwise, the proposed edit is discarded and the previous skill space is
restored. This validation-and-rollback mechanism acts as a trust-region-style constraint
in skill space, preventing unstable edits caused by noisy critiques,
or overfitting to recent trajectories.


\begin{table*}[t]
\centering
\caption{Main results across HotpotQA, LoCoMo, and GAIA.} 
\label{tab:main}

\begin{subtable}{\textwidth}
\centering
\caption{Multi-Hop Reasoning: HotpotQA}
\resizebox{\textwidth}{!}{
\begin{tabular}{lccccccccc}
\toprule
\textbf{Method} & IO & CoT & CoT-SC & MedPrompt & MultiPersona & Self-Refine & ADAS & \textbf{MASkills} \\
\midrule
F1 & 68.1 & 67.9 & 68.9 & 68.3 & 69.2 & 60.8 & 64.5 & \textbf{76.3} \\
\bottomrule
\end{tabular}
}
\end{subtable}

\vspace{0.8em}

\begin{subtable}[t]{0.55\textwidth}
\centering
\caption{Long-horizon Memory: LoCoMo}
\resizebox{\linewidth}{!}{%
\begin{tabular}{lcccc}
\toprule
\textbf{Method} & \textbf{SH-F1} & \textbf{SH-BLEU} & \textbf{MH-F1} & \textbf{MH-BLEU} \\
\midrule
MemoryBank & 5.00 & 4.77 & 5.56 & 5.94 \\
ReadAgent & 9.15 & 6.48 & 5.31 & 5.12 \\
LoCoMo & 25.02 & 19.75 & 12.04 & 11.16 \\
MemGPT & 26.65 & 17.72 & 9.15 & 7.44 \\
\midrule
\textbf{MASkills} & \textbf{27.61} & \textbf{21.30} & \textbf{17.22} & \textbf{12.87} \\
\bottomrule
\end{tabular}
}
\end{subtable}
\hfill
\begin{subtable}[t]{0.42\textwidth}
\centering
\caption{General AI Assistants Benchmark}
\resizebox{\linewidth}{!}{%
\begin{tabular}{lcccc}
\toprule
\textbf{Framework} & \textbf{L1} & \textbf{L2} & \textbf{L3} & \textbf{Avg.} \\
\midrule
Base & 12.8 & 3.8 & 0.0 & 6.8 \\
Search-o1 & 23.1 & 17.3 & 0.0 & 17.5 \\
Vanilla ReAct & 28.2 & 15.3 & 0.0 & 18.4 \\
R1-Searcher & 28.2 & 19.2 & \textbf{8.3} & 20.4 \\
\textbf{MASkills} & \textbf{35.3} & \textbf{22.6} & 0.0 & \textbf{23.3} \\
\bottomrule
\end{tabular}
}
\end{subtable}

\end{table*}

\begin{figure*}[htbp]
    \centering
    \includegraphics[width=1.0\textwidth]{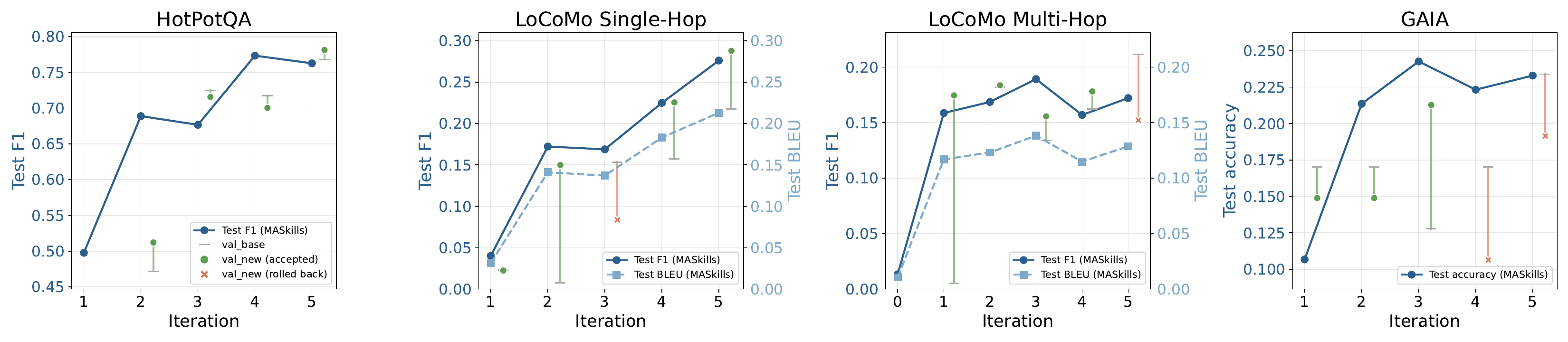}
    \caption{Training curves of MASkills across benchmarks with validation-based skill update decisions.}
    \label{fig:training_curve}
\end{figure*}

\section{Experiments}
\label{sec:experiments}

We organize the empirical evaluation around three research questions (RQ1--RQ3).

\noindent$\bullet$~\textbf{RQ1 (Task Performance):}
Does continual skill optimization improve the performance of multi-agent LLM systems across diverse tasks, and which components contribute most to the improvements?

\noindent$\bullet$~\textbf{RQ2 (Skill Quality and Transferability):}
Do continually optimized skills become more reusable, transferable, and coordination-effective over time?

\noindent$\bullet$~\textbf{RQ3 (Robustness and Generalization):}
Is MASkills robust across different multi-agent coordination topologies and LLM backbones?

\subsection{Experimental Setup}

We instantiate all tasks as cooperative Dec-POMDP environments in which a team of specialized LLM agents interacts under decentralized execution and a shared team-level objective. Each agent maintains a local observation space and executes a role-specialized policy conditioned on its observation history and skill library. To enable coordinated problem solving, agents are assigned complementary functional responsibilities, such as information retrieval, verification, planning, memory tracking, tool use, and decision making. Agents communicate through decentralized peer-to-peer interaction.

For continual skill optimization, we use GPT-5.1 as the optimizer backbone responsible for skill-conditioned credit assignment, trajectory aggregation, skill refinement, induction, consolidation, and pruning. Following prior work~\cite{chhikara2025mem0, zhang2025aflow, webdancer}, we use GPT-4o-mini (HotpotQA and LoCoMo) and Qwen2.5-7B (GAIA) as the primary actor backbones for agent execution.

\subsection{Benchmarks}

\paragraph{Tasks.} We instantiate experiments on three benchmarks: (1) \textbf{Multi-hop Reasoning}: HotpotQA~\cite{yang2018hotpotqa}, a multi-document question answering benchmark requiring compositional reasoning across supporting evidence. (2) \textbf{Long-horizon dialogue memory}: LoCoMo~\cite{maharana2024locomo}, a very-long-term multi-session conversation benchmark designed to evaluate persistent conversational memory and temporal reasoning. (3) \textbf{General AI Assistants Benchmark (GAIA)}~\cite{mialon2024gaia} is a benchmark designed to evaluate general AI agents on real-world tasks, with a focus on reasoning, multimodal understanding, web browsing, and tool-use capabilities.

\paragraph{Metrics.}
We report both benchmark-specific task metrics and skill-evolution diagnostics. For \textit{HotpotQA}, we follow the standard evaluation protocol and report answer-level F1. For \textit{LoCoMo}, we report both F1 and BLEU across the four question categories defined by the benchmark (multi-hop, temporal, open-domain, and single-hop memory queries). For \textit{GAIA}, we report averaged success rate across various tasks.

\subsection{Main Results (RQ1)}

Table~\ref{tab:main} summarizes the main results on HotpotQA, LoCoMo, and GAIA. Overall, MASkills achieves strong and consistent performance across a diverse set of reasoning, long-context memory, and agentic problem-solving benchmarks, demonstrating the effectiveness of skill optimization in multi-agent systems. On HotpotQA, MASkills outperforms existing prompting-based and multi-agent reasoning baselines, indicating stronger multi-hop reasoning and evidence integration capabilities. On LoCoMo, MASkills consistently improves long-context memory retrieval and response quality across both single-hop and multi-hop settings, suggesting better temporal reasoning and memory coordination over extended interactions. On GAIA, MASkills achieves the strongest overall performance among compared agent frameworks, demonstrating improved planning, decomposition, and tool-use abilities in open-ended environments. The training dynamics of MASkills are shown in Figure~\ref{fig:training_curve}, showing how skill optimization contributes to performance improvement over time.

\begin{figure*}[t]
    \centering
    \includegraphics[width=1.0\textwidth]{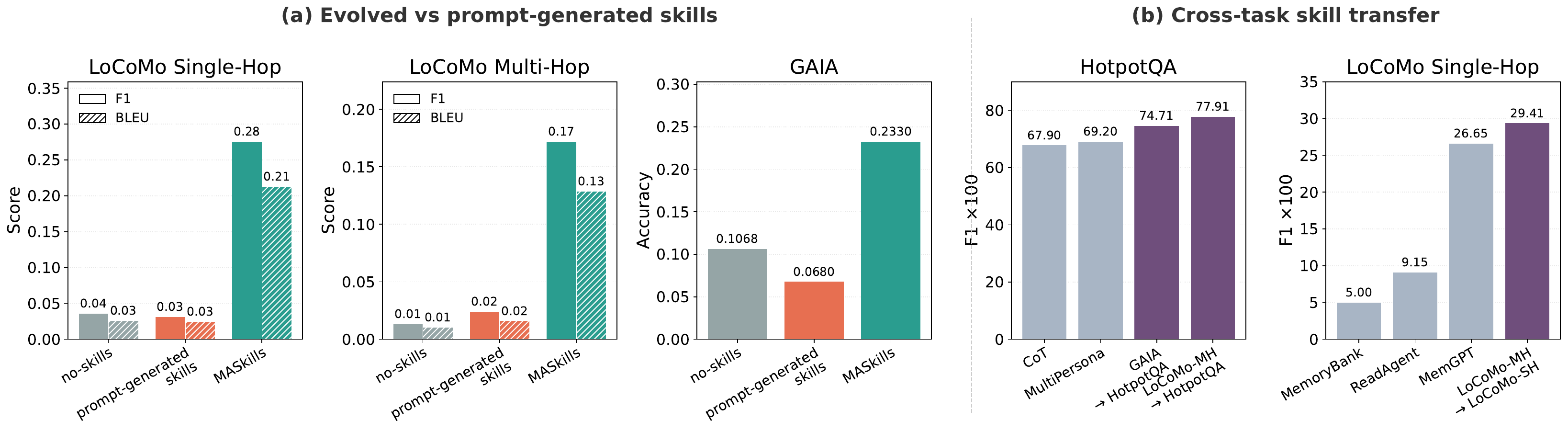}
    \caption{Skill quality and transferability analysis of MASkills. (a) Comparison between evolved skills, prompt-generated skills, and no-skill baselines on LoCoMo and GAIA. (b) Cross-task skill transfer results show that skills learned in one environment generalize effectively to unseen target tasks, indicating that MASkills acquires reusable procedural abstractions rather than task-specific prompting heuristics.}
    \label{fig:transfer}
\end{figure*}

\subsection{Skill Quality and Transferability (RQ2)}

We next evaluate whether continually optimized skills become more reusable, transferable, and coordination-effective over time. 
To this end, we analyze both the intrinsic quality of evolved skills and their ability to generalize across benchmarks.

Figure~\ref{fig:transfer}(a) compares MASkills against no-skill baselines and prompt-generated skills across LoCoMo and GAIA. 
We observe that prompt-generated skills provide only marginal improvements over directly prompting the agents, while continually optimized skills learned by MASkills lead to substantial gains across all evaluated settings. 
In particular, MASkills significantly improves both F1 and BLEU on LoCoMo single-hop and multi-hop memory tasks, while also achieving the strongest performance on GAIA. 
These results suggest that iterative skill refinement and trajectory-driven optimization produce substantially higher-quality procedural abstractions than one-shot prompt-generated skills.

Figure~\ref{fig:transfer}(b) further evaluates cross-task skill transfer. 
We transfer skill libraries learned from one environment into previously unseen target tasks without additional optimization. 
Transferred skills consistently improve downstream performance across all evaluated source-target pairs. 
For example, skills optimized on GAIA improve HotpotQA reasoning performance beyond strong prompting-based baselines such as CoT and MultiPersona, while LoCoMo-derived skills improve downstream long-horizon memory reasoning. 
These findings suggest that MASkills captures some reusable behavioral patterns.

\begin{figure}[!t]
    \centering

    \begin{minipage}[t]{0.40\linewidth}
        \centering
        \vspace{0pt}
        \includegraphics[width=\linewidth]{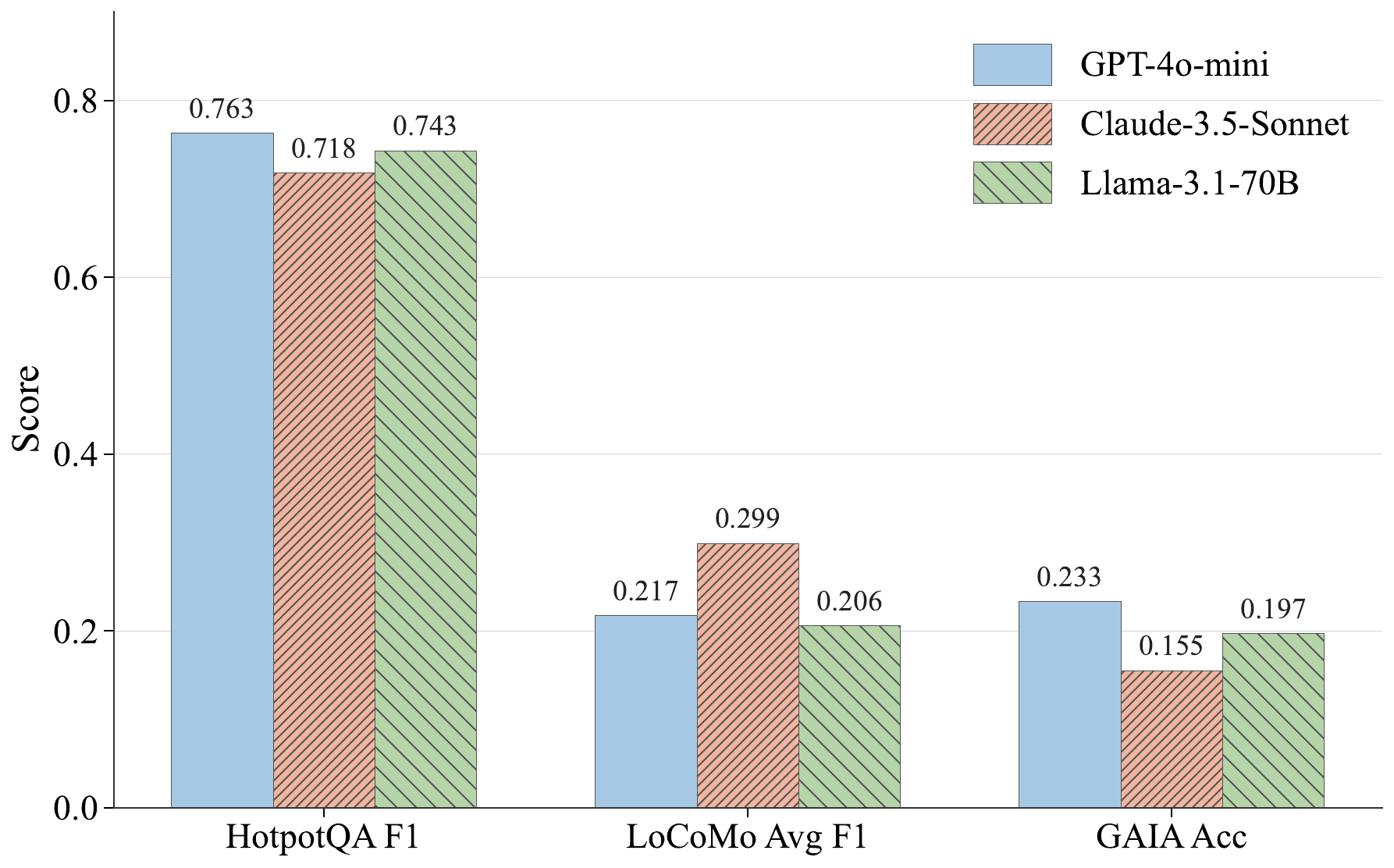}
        \caption{Backbone comparison across different underlying language models.}
        \label{fig:backbone_comparison}
    \end{minipage}
    \hfill
    \begin{minipage}[t]{0.55\linewidth}
        \centering
        \vspace{6mm}
        \small
        \resizebox{\linewidth}{!}{
        \begin{tabular}{lcc}
        \toprule
        Method & LoCoMo-MH & GAIA \\
        \midrule
        MASkills (Full) & \textbf{17.2} & \textbf{23.3} \\
        \quad w/o Skill Credit assignment & 14.2 & 17.1 \\
        \quad w/o Momentum Smoothing & 16.4 & 21.9 \\
        \quad w/o Validation Rollback & 6.6 & 13.5 \\
        \quad w/o Consolidation / Pruning & 13.9 & 13.0 \\
        \bottomrule
        \end{tabular}
        }
        \captionof{table}{Ablation Study}
        \label{tab:ablation}
    \end{minipage}

\end{figure}

\subsection{Robustness and Generalization (RQ3)}

We next evaluate whether MASkills generalizes across different coordination structures and underlying language model backbones.

\paragraph{Topology Robustness.}
We instantiate MASkills under centralized, decentralized peer, and hierarchical structures while keeping the optimization pipeline unchanged. Table~\ref{tab:topology} shows that MASkills maintains competitive performance across different coordination topologies, suggesting that the proposed skill optimization framework is not tightly coupled with a specific orchestration structure.

Interestingly, the optimal topology is highly task-dependent. On HotpotQA and GAIA, decentralized peer coordination achieves the strongest performance. These tasks benefit from diverse exploration, independent tool or retrieval trajectories. In contrast, centralized coordination performs best on LoCoMo, particularly on multi-hop memory retrieval. Unlike retrieval-centric tasks, LoCoMo requires maintaining globally consistent long-context memory representations, resolving temporal dependencies, and integrating information across multiple conversational sessions. Centralized coordination provides a shared global state that reduces memory fragmentation and inconsistency across agents, leading to more coherent long-horizon reasoning and generation quality. Hierarchical coordination generally yields intermediate performance, balancing structured information aggregation with partial decentralization. 

\paragraph{Backbone Generalization.}
We further evaluate MASkills across multiple LLM backbones, including both proprietary and open-source models. Figure~\ref{fig:backbone_comparison} shows that MASkills maintains competitive performance across different model families on HotpotQA, LoCoMo, and GAIA. While the relative strengths of the backbones vary across tasks, the overall trends remain consistent: stronger reasoning-oriented models perform better on long-context and agentic tasks, whereas open-weight models still achieve solid performance under the MASkills framework.

\begin{table*}[t]
\centering
\small
\resizebox{\textwidth}{!}{
\begin{tabular}{lcccccccc}
\toprule
Topology & HotpotQA F1 & LoCoMo-SH F1 & LoCoMo-SH BLEU & LoCoMo-MH F1 & LoCoMo-MH BLEU & GAIA-L1 &
GAIA-L2 & GAIA-Avg \\
\midrule
Centralized           & 72.46          & \textbf{27.68} & 20.76          & \textbf{22.27} &
\textbf{17.70} & 17.65          & 11.54          & 11.76          \\
Hierarchical          & 71.86          & 27.55          & \textbf{20.85} & 22.03          & 17.62
         & 20.59          & 11.54          & 12.75          \\
Decentralized Peer    & \textbf{76.30} & 27.61          & 21.30          & 17.22          & 12.87
         & \textbf{35.30} & \textbf{22.60} & \textbf{23.30} \\
\bottomrule
\end{tabular}
}
\caption{Performance under different coordination topologies.}
\label{tab:topology}
\end{table*}

\subsection{Ablation Study}

To understand the contribution of different components in MASkills, we conduct ablation studies over skill-conditioned credit assignment, momentum smoothing, validation rollback, and consolidation/pruning. Table~\ref{tab:ablation} shows that removing any component leads to performance degradation, indicating that all modules contribute to stable continual skill optimization. Among them, validation rollback has the largest impact, suggesting that reverting unstable updates is critical for preventing performance collapse during iterative optimization. Skill-conditioned credit assignment also plays an important role by improving the attribution of successful behaviors to specific skills. In addition, momentum smoothing and consolidation/pruning provide consistent gains by stabilizing optimization and reducing redundant or low-quality skills.

\section{Conclusion}
\label{sec:conclusion}

We presented MASkills, a continual learning framework that optimizes multi-agent LLM systems in skill space. Through skill-conditioned credit assignment and evolution, MASkills enables agents to continually refine and reuse skills from interaction experience. Experiments on various benchmarks show that MASkills consistently improves memory and agentic tasks, highlighting the potential of continual skill optimization for scalable and adaptive multi-agent systems.

\section*{Acknowledgement}
The work was partially supported by Cisco Faculty Award, Amazon Research Award, and NSF awards \#2442477, \#2550203, \#2616632, \#2623317, \#2536297, and \#2613637. The views and conclusions in this paper should not be interpreted as representing any funding agencies.

\section*{Limitations}

Although MASkills demonstrates strong performance across multiple benchmarks, several limitations remain. First, our current experiments mainly focus on cooperative settings with relatively fixed agent roles and communication topologies, leaving more dynamic, adversarial, and open-world environments underexplored. Future work could extend MASkills to adaptive organizational structures, competitive multi-agent games, and large-scale decentralized coordination scenarios. Second, as the skill library continually expands during optimization, scalability challenges may emerge in skill retrieval, consolidation, and coordination efficiency. Future research may investigate hierarchical skill organization, retrieval compression, and lifelong learning mechanisms to support efficient long-term continual skill evolution at scale.

\section*{Ethical Considerations}

MASkills aims to improve multi-agent LLM systems by continually refining reusable skill artifacts from interaction experience. While this can enhance agent performance and adaptability, it also introduces several ethical considerations. First, automatically evolved skills may amplify undesirable behaviors inherited from the underlying LLMs, including factual errors, social biases, unsafe tool-use patterns, or overconfident reasoning. To mitigate this risk, MASkills incorporates held-out validation, rollback, consolidation, and pruning mechanisms before committing skill updates; however, these mechanisms should not be viewed as a substitute for human oversight in high-stakes applications. Second, because the framework learns from interaction trajectories, deployments involving user data should follow strict privacy practices, including data minimization, anonymization, access control, and compliance with the licenses and usage terms of the underlying datasets and tools. Third, improved multi-agent coordination and tool-use ability could be misused for harmful automation if applied without appropriate safeguards. We therefore recommend restricting MASkills to benign use cases, monitoring evolved skills for unsafe or deceptive behavior, and maintaining transparent audit logs of skill modifications. Our experiments are conducted on established research benchmarks, and we do not intend the framework to be used for decision-making in sensitive domains without additional safety evaluation, domain-specific validation, and human accountability.

\section*{AI usage}
During the preparation of this work, the authors employed LLMs to assist with code implementation and to refine the manuscript's grammatical clarity and stylistic flow. All final content, technical contributions, and experimental analyses were rigorously reviewed and verified by the authors.



\end{document}